\pdfoutput=1
\documentclass[11pt]{article}

\usepackage[preprint]{acl}

\usepackage{times}
\usepackage{latexsym}
\usepackage{amsmath}
\usepackage[T1]{fontenc}
\usepackage[utf8]{inputenc}

\usepackage{microtype}

\usepackage{inconsolata}

\usepackage{graphicx}
\usepackage{tablefootnote}

\usepackage{listings}
\usepackage{tcolorbox}
\usepackage{lipsum}

\usepackage{booktabs}
\usepackage{multirow}
\title{DynamiCare: A Dynamic Multi-Agent Framework for Interactive and Open-Ended Medical Decision-Making}

\author{
 \textbf{Tianqi Shang\textsuperscript{1}},
 \textbf{Weiqing He\textsuperscript{1}},
 \textbf{Charles Zheng\textsuperscript{1}},
 \textbf{Lingyao Li\textsuperscript{2}},
\\
 \textbf{Li Shen\textsuperscript{1}},
 \textbf{Bingxin Zhao\textsuperscript{1,*}}
\\
\\
 \textsuperscript{1}University of Pennsylvania,
 \textsuperscript{2}University of South Florida
\\
 \small{
 \texttt{tianqi.shang@pennmedicine.upenn.edu, weiqingh@sas.upenn.edu, czbw@seas.upenn.edu}
 }
 \\
 \small{
 \texttt{lingyaol@usf.edu, li.shen@pennmedicine.upenn.edu, bxzhao@wharton.upenn.edu}
 }
\\
 \small{
   \textsuperscript{*}\textbf{Corresponding Author}
 }
}

\begin{document}
\maketitle
\begin{abstract}
The rise of Large Language Models (LLMs) has enabled the development of specialized AI agents with domain-specific reasoning and interaction capabilities, particularly in healthcare. While recent frameworks simulate medical decision-making, they largely focus on single-turn tasks where a doctor agent receives full case information upfront—diverging from the real-world diagnostic process, which is inherently uncertain, interactive, and iterative. In this paper, we introduce \textbf{MIMIC-Patient}, a structured dataset built from the MIMIC-III electronic health records (EHRs), designed to support dynamic, patient-level simulations. Building on this, we propose \textbf{DynamiCare}, a novel \textbf{dynamic multi-agent framework} that models clinical diagnosis as a multi-round, interactive loop, where a team of specialist agents iteratively queries the patient system, integrates new information, and dynamically adapts its composition and strategy. We demonstrate the feasibility and effectiveness of DynamiCare through extensive experiments, establishing the \textbf{first benchmark} for dynamic clinical decision-making with LLM-powered agents.
\end{abstract}  

\section{Introduction}

\begin{figure*}[t]
    \centering
    \includegraphics[width=\linewidth]{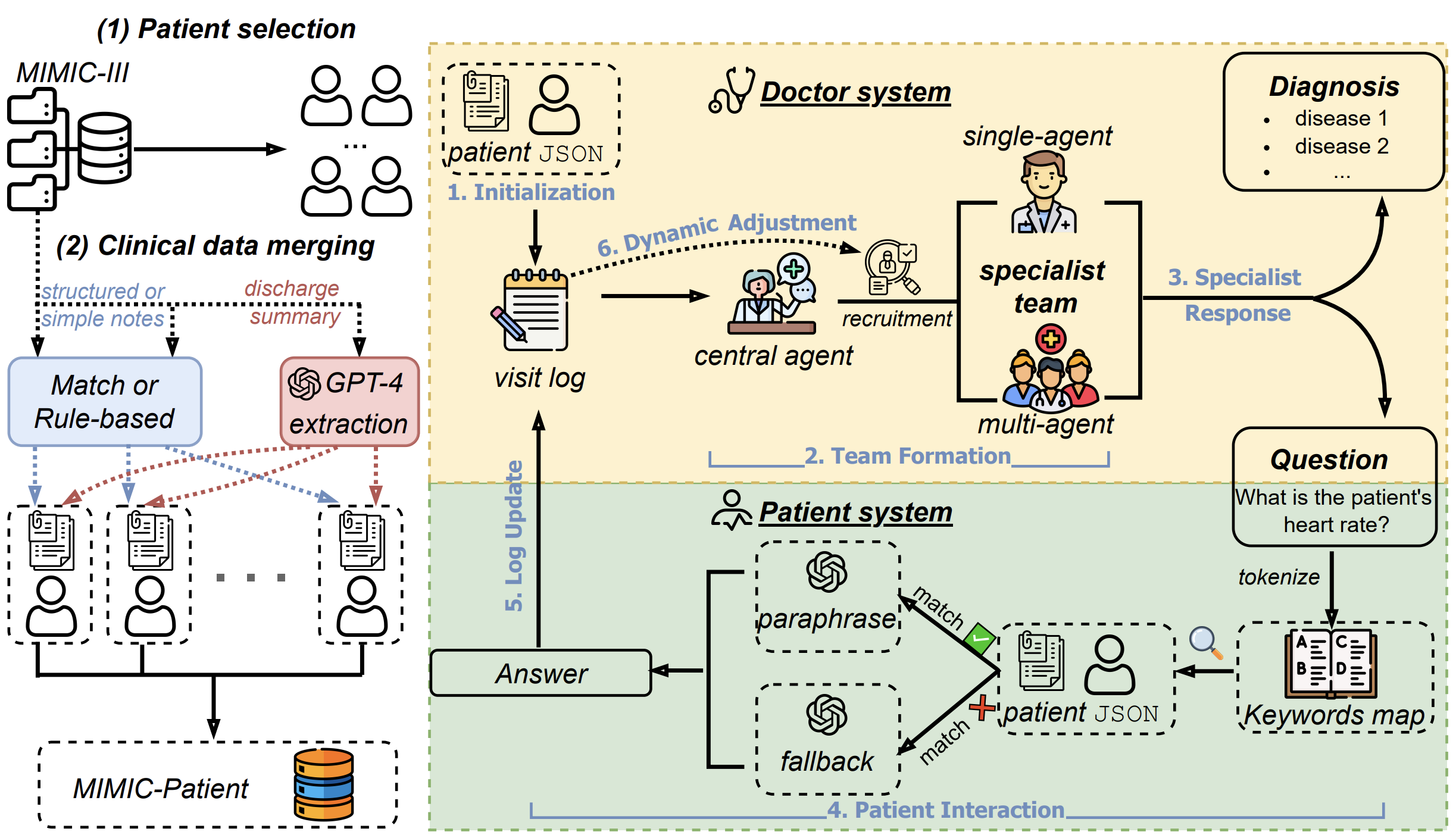}
    \caption{Illustration of the MIMIC-Patient dataset and DynamiCare framework. \textbf{Left}: the construction process of MIMIC-Patient. \textbf{Right}: The DynamiCare framework, which consists of a Patient System and a Doctor System, operating in a six-step loop: 1) initialization; 2) team formation; 3) specialist response; 4) patient interaction; 5) log update; 6) dynamic adjustment.}
    \label{fig:intro}
\end{figure*}
 
The advent of Large Language Models (LLMs) has laid the foundation for developing specialized AI agents capable of reasoning and interaction tailored to applications in the healthcare domain~\cite{clusmann2023future, kim2024health, saab2024capabilities, truhn2024large, zhou2023survey}. Recent works have leveraged LLMs to simulate medical decision-making~\cite{kim2024mdagents, li2024mediq, fan2024ai, jin2024agentmd, li2024agent, tang2023medagents}, complemented by the creation of diverse datasets~\cite{jin2019pubmedqa, jin2021disease, pal2022medmcqa, chen2025benchmarking} designed to mimic real-world medical scenarios and facilitate systematic evaluations.

However, most current AI agents focus on single-turn question-answering tasks~\cite{kim2024mdagents, li2023camel, wu2023autogen, du2023improving, wang2022self, tang2023medagents, chen2023reconcile}. In these scenarios, an LLM-based agent often receives a complete description of an illness at the beginning of the study and is expected to immediately provide a diagnosis. This approach deviates from real clinical practice in two significant ways. First, \textit{the ``question'' data does not reflect the real diagnostic process}, where patients rarely present complete conditions at the start. Second,\textit{medical diagnosis involves interactive and iterative exchanges}, where healthcare providers progressively elicit relevant details (e.g., symptoms, history, and lab test results) through multiple rounds of interactions. 

While some recent studies~\cite{li2024mediq, hu2024uncertainty, schmidgall2024agentclinic} have explored interactive diagnostic frameworks with incomplete initial information, they often lack true dynamism—\textit{the ability to adapt the composition and behavior of the agent team based on newly acquired information}. In contrast, real clinical environments necessitate continuous, context-aware adjustments, including modifications to the healthcare team structure in response to evolving patient needs and clinical complexity.

To address these limitations, we propose \textbf{MIMIC-Patient} (the left part of Figure~\ref{fig:intro}), a patient-level dataset that compiles diverse medical information for each patient based on the MIMIC-III Clinical database~\cite{johnson2016mimic}, and \textbf{DynamiCare} (the right part of Figure~\ref{fig:intro}), a dynamic, interactive framework for simulating clinical decision-making. DynamiCare consists of a Patient System and a Doctor System, operating in a six-step loop:
1) Initialization: create a visit log using basic patient information from MIMIC-Patient; 2) Team Formation: a central agent recruits specialist agents based on the visit log; 3) Specialist Response: the team generates a diagnosis or follow-up question; 4) Patient Interaction: the Patient System answers the question using the patient record; 5) Log Update: the Q\&A pair is added to the visit log; 6) Dynamic Adjustment: the central agent updates the team based on the new information, and the loop continues until a diagnosis is made or reaches the interaction round limit. 

Our contributions are summarized as follows:
 \begin{itemize}
     \item We establish the \textbf{MIMIC-Patient} benchmark, a patient-centric benchmark dataset derived from MIMIC-III, which structures diverse medical information to support interactive and open-ended decision-making tasks.
     \item We introduce \textbf{DynamiCare}, a \textbf{novel multi-agent framework} to model clinical reasoning as a dynamic, interactive process that can adapt its structure and strategy based on newly acquired patient information.
     \item We conduct extensive experiments using DynamiCare on MIMIC-Patient, establishing a \textbf{benchmark} that set a foundation for future research on dynamic medical agents.
 \end{itemize}

\section{Related Works}

\paragraph{Clinical Dataset} Recent works~\cite{singhal2023large, singhal2025toward, nori2023capabilities, lievin2024can} have produced a wide range of benchmark datasets to evaluate the performance of LLMs in the clinical domain. Notable examples include MedQA~\cite{jin2021disease}, PubMedQA~\cite{jin2019pubmedqa}, and MedMCQA~\cite{pal2022medmcqa}, which feature question-answer pairs derived from medical licensing exams or biomedical literature. 
Additionally, several works~\cite{pampari2018emrqa, fan2019annotating, kweon2024ehrnoteqa} have explored electronic health record (EHR) datasets. For example, \citet{kweon2024ehrnoteqa} introduced EHRnoteQA, a QA dataset derived from discharge summaries in MIMIC-IV~\cite{johnson2023mimic}. Although valuable for assessing the factual medical knowledge of LLMs, these datasets only include pre-defined questions and answers without the ability to support the dynamic, interactive modeling of real clinical encounters. To address this, \citet{li2024mediq} proposed MedIQ, an interactive benchmark for medical evaluation. However, the construction is based on MedQA and Craft-MD~\cite{johri2024craft}, which lacks the complexity of real-world clinical scenarios. In practice, clinical tasks often involve extracting, synthesizing, and reasoning over heterogeneous patient information—such as clinical notes, laboratory results, and timelines of care—which these benchmarks do not often capture.

% While valuable for assessing the factual medical knowledge of LLMs, these datasets fall short in representing the complexity of real-world clinical scenarios. In practice, clinical tasks often involve extracting, synthesizing, and reasoning over heterogeneous patient information—such as clinical notes, laboratory results, and timelines of care—which these benchmarks do not capture. 

% To address this gap, several works~\cite{pampari2018emrqa, fan2019annotating, kweon2024ehrnoteqa} have explored electronic health record (EHR)-based datasets. Recently, \citet{kweon2024ehrnoteqa} introduced EHRnoteQA, a QA dataset derived from discharge summaries in MIMIC-IV, aiming to test model performance on more realistic, document-grounded tasks. However, these datasets remain static, offering pre-defined questions and answers without modeling the dynamic, interactive nature of real clinical encounters.

\paragraph{LLM Agents in Medical Decision Making}
A growing body of research~\cite{zhou2023survey, wang2024survey} has proposed frameworks that leverage single or multiple LLM agents to support medical decision-making. These approaches typically rely on prompt engineering to guide LLMs in completing clinical tasks. Notable paradigms include role-playing~\cite{li2023camel, wu2023autogen}, debate~\cite{du2023improving, liang2023encouraging}, voting~\cite{wang2022self}, multi-disciplinary collaboration~\cite{tang2023medagents}, and group discussion~\cite{chen2023reconcile}. While these agentic frameworks have demonstrated performance improvements in specific settings, they are often built around a fixed and pre-defined set of agents, making them inherently static.
To address this rigidity, \citet{kim2024mdagents} introduced MDAgents, a more adaptive framework that employs a complexity assessment to determine the appropriate number of agents for a given task. However, MDAgents can fall short of capturing the dynamic nature of real-world clinical workflows, where the composition and involvement of specialist agents should evolve over the course of multi-turn interactions between clinicians and patients.

% Our work comprises two main components: the construction of MIMIC-Patient, followed by a description of DynamiCare. In this section, we will first detail the development of MIMIC-Patient, and then present DynamiCare.

\section{MIMIC-Patient}
\label{sec: data}
We build MIMIC-Patient, a dataset derived from the MIMIC-III database~\cite{johnson2016mimic}. MIMIC-III is a large, publicly available dataset containing de-identified clinical data from over 40,000 patients admitted to critical care units at the Beth Israel Deaconess Medical Center between 2001 and 2012. However, it can be extremely challenging for our dynamic scenarios given its inherent complexity of clinical records. For instance, individual patient admission may include hundreds of diagnoses—far exceeding the input capacity of current LLMs. Moreover, the clinical data are distributed across multiple relational tables, complicating effective integration and interpretation. To address these challenges, we adopt a two-stage data processing approach (the left part of Figure~\ref{fig:intro}) to construct a more structured and analysis-friendly dataset called MIMIC-Patient.

%%%%table: mimic-patient%%%%%
\begin{table}[t]
\centering
\begin{tabular}{ll}
\toprule
\textbf{Data Type} & \textbf{Clinical Info} \\
\midrule
\multirow{7}{*}{Structured Data} 
 & Admission Info \\
 & Demographics \\
 & Diagnoses \\
 & Prescription \\
 & Procedure~\tablefootnote{Clinical procedures performed, with codes and timestamps.} \\
 & Chart Data \\
 & Lab Data \\
\midrule
\multirow{3}{*}{Semi-Structured Text} 
 & ECG reports~\tablefootnote{Electrocardiogram reports and interpretations.} \\
 & Echo reports~\tablefootnote{Echocardiogram results, including measurements like ejection
fraction.} \\
 & Radiology reports\\
\midrule
Unstructured Text & Discharge Summary \\
\bottomrule
\end{tabular}
\caption{Clinical information included in the patient \texttt{JSON}.\tablefootnote{More details of the database can be found at: https://mimic.mit.edu/docs/iii/tables/}}
\label{table: mimic-patient}
\end{table}
%%%%%%%%%%%%%%

\paragraph{Data selection} MIMIC-III contains 58,976 hospital admissions, each representing a unique hospital stay with associated clinical data. We begin by selecting admissions that met all three criteria:
\begin{itemize}
    \item The admission has fewer than five diagnosed diseases. In the MIMIC-III dataset, some admissions have a large number of multimorbidities, which can introduce excessive complexity and noise.
    \item The admission is neither for a newborn nor deceased. Newborns often follow different clinical pathways and have distinct data structures, while deceased patients may have incomplete or atypical medical records that could bias the results.
    \item The admission has sufficient clinical data available, i.e., at least containing all structured data and a discharge summary, to ensure that both the doctor and patient agents have access to meaningful and informative input for decision-making.
\end{itemize}
After filtering for valid admissions, 2,597 remained. Note that a single patient may have multiple admissions. To ensure a one-to-one mapping between patients and admissions, we retain only one admission per patient, resulting in 2,452 unique admissions. From this subset, we then randomly select 500 admissions. Since each admission corresponds to a unique patient in a one-to-one manner, we refer to our dataset as patient-level and use the term ``patient'' interchangeably with the corresponding admission in the following descriptions.

\paragraph{Clinical data merging}
In the MIMIC-III database, clinical data are distributed across multiple tables. To centralize and structure this information, we extract and merge the relevant data for each patient into a single \texttt{JSON} file (see examples in Appendix~\ref{app: patient}).
For structured data (e.g., charted observations), we directly match and integrate the entries based on patient IDs. For unstructured data (e.g., free-text clinical notes), we employ two extraction methods. First, we develop a \textit{rule-based} approach to extract information from semi-structured text—such as the \texttt{Impression} and \texttt{Findings} sections in radiology reports—by identifying and normalizing known section headers. Second, for more complex unstructured texts like discharge summaries, we leverage GPT-4~\cite{achiam2023gpt} to parse the content and generate a structured \texttt{JSON} representation.

As a result, the final dataset comprises 500 \texttt{JSON} files—one for each patient—containing a variety of clinical records, from demographic details to lab results. A complete list of the included clinical information is provided in Table~\ref{table: mimic-patient}.

\section{DynamiCare}
\label{sec: Agent}
To enable interactive medical decision-making, DynamiCare (the right part of Figure~\ref{fig:intro}) is designed to comprise two components: the \textbf{Patient System} and the \textbf{Doctor System}. At a high level, the Patient System responds to queries from the Doctor System, which in turn updates its strategy based on the received information. In this section, we first introduce the Patient System, followed by the Doctor System. We then describe the overall workflow and explain how DynamiCare facilitates dynamic interactions.
\begin{figure*}[t!]
    \centering
    \includegraphics[width=\linewidth]{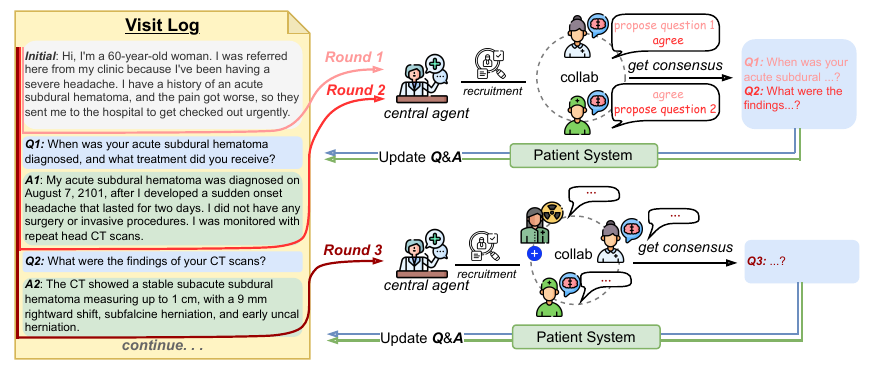}
    \caption{Illustrative example of DynamiCare with a multi-specialist team. The figure presents three rounds of interaction, where in each round the specialist team poses a question to the patient. In Rounds 1 and 2, the team consists of a neurologist and a neurosurgeon. Notably, in Round 3, the Central Agent recruits an additional specialist—a radiologist—to assist with analyzing the patient’s CT scans.}
    \label{fig:detail}
\end{figure*}

\subsection{Patient System}
To enable interactive querying of patient information by a doctor agent (e.g., an LLM), we build a Patient System capable of responding to natural language questions using structured patient records. Importantly, doctor queries vary in complexity—some map directly to specific data fields, while others require interpretation and integration across multiple record types. In order to ensure the reliability of its responses and minimize LLM hallucinations, we propose a two-stage answering process (the green part of Figure~\ref{fig:intro}) that combines rule-based keyword matching with fallback LLM inference.

For each query, the Patient System first extracts relevant keywords using regular expression-based tokenization. These keywords are then mapped to specific sections of the patient’s structured \texttt{JSON} file—such as demographics, prescriptions, or lab results—based on a predefined keyword-to-section mapping dictionary (example in Appendix~\ref{app:key-words}). Once the relevant section is identified, the Patient System retrieves the corresponding information and uses GPT-4.1~\cite{openai2025gpt41} to formulate a response in the patient's voice. 

To handle cases where no section is matched or the retrieved information is ambiguous or missing, the system falls back to direct LLM inference. In such scenarios, the patient \texttt{JSON} (de-identified, removing admission and demographics) is provided as context to GPT-4.1, which then generates an answer based on the broader clinical context. 

The generated answer is then passed to the Doctor System to support real-time adaptation of its composition and strategy.

\subsection{Doctor System}
\label{sec: doctor system}
% 1. Central Agent: acts as a central medical coordinator(someone senior and comprehensive) overseeing a diagnostic team. It will analyze the current visit log(containing Q&As from previous rounds of interaction) and decide whether to update the specialist team(one or more specialists, each with a distinct role) (reason for update might be new information arrives so a certain area specialist is needed/no longer needed)
% 2. Specialist Team(one or more specialists, each with a distinct role): based on the current patient case, decide whether to make a diagnosis or request more information with a new question. The specialist team has two situations with slight different workflow:
% 2.1 Single agent: analyze current visit log -> decide diagnosis confidence -> if high: generate diagnosis; if low: generate a question
% 2.2 Multi-agent: analyze current visit log -> All agents propose a response (question or diagnosis) with a confidence score -> every agent votes on every other agent's response. -> The first response to get enough agreement votes (≥ threshold) is accepted. -> If no agreement is reached after voting on all responses - return the one with highest confidence.
Previous studies~\cite{kim2024mdagents, gilboy2012emergency, christ2010modern, wuerz2000reliability} have shown that flexible adaptation of specialist teams to varying patient conditions is both effective and promising for multi-agent system design. Moreover, as the doctor agent actively guides the interaction to gather additional information, the patient’s condition representation may evolve. This highlights the need for a dynamic Doctor System capable of adjusting its composition in real time.

\paragraph{Dynamic Specialist Team} In DynamiCare, the Doctor System (the yellow section of Figure~\ref{fig:intro}) incorporates a \textbf{Central Agent} that acts as a senior medical coordinator, dynamically managing a \textbf{Specialist Team} based on the evolving context of each patient case. The Central Agent continuously reviews the visit log—which captures previous rounds of question-answer interactions—and decides whether to update the composition of the Specialist Team. This dynamic process ensures that the team remains aligned with the most current information, adding or removing specialists as needed. 

\paragraph{Collaborative Decision Protocols} For simpler cases, the Central Agent may assign a single specialist, while more complex scenarios may involve multiple specialists with complementary expertise. The specialist team then decides whether to issue a diagnosis or request additional information. \textit{In the single-specialist setting}, the agent analyzes the visit log, evaluates diagnostic confidence, and either provides a diagnosis (if confidence is high) or asks a follow-up question (if confidence is low). \textit{In the multi-specialist setting}, each agent independently proposes a response (diagnosis or question) along with a confidence score. Specialists then vote on each other's proposals; the first response to meet a predefined agreement threshold is accepted. If no consensus is reached, the response with the highest confidence is selected. This adaptable framework ensures that diagnostic reasoning is both context-aware and continuously updated in response to new patient information.

\subsection{Dynamic Interaction Workflow}

The interaction between the Patient and Doctor Systems follows a dynamic, iterative workflow that enables real-time information gathering and diagnosis. Figure~\ref{fig:detail} gives a detailed illustration with a concrete example for this workflow. The process proceeds in six steps:
\begin{enumerate}
    \item \textbf{Initialize Visit Log}: A \textit{visit log} is created using the patient's basic information, which contains demographics, basic relevant symptoms, and the reason for the visit.
    \item \textbf{Specialist Team Formation}: The \textit{Central Agent} receives the initial visit log. Based on this input, the Central Agent initializes an appropriate \textit{Specialist Team} tailored to the current case complexity.
    \item \textbf{Specialist Response Generation}: The Specialist Team analyzes the current visit log and, following the protocols described in Section~\ref{sec: doctor system}, generates a response. This response can be either a follow-up question or a diagnostic conclusion. \label{workflow: specialist response}
    \item \textbf{Patient Interaction}: If the response is a diagnosis, it is treated as the final decision, and the interaction ends. If the response is a question, the \textit{Patient System} is queried to retrieve an appropriate answer.

    \item \textbf{Update Visit Log}: The current question-answer pair is appended to the visit log to maintain a comprehensive record of the evolving case.
    \item \textbf{Dynamic Adjustment}: The Central Agent re-evaluates the current visit log and the performance of the existing specialist team. Based on new information or shifting diagnostic needs, it may revise the team composition before returning to Step~\ref{workflow: specialist response} for next-round analysis.
\end{enumerate}
Additionally, to ensure termination, a manual stopping criterion is defined: if the system reaches a predefined maximum number of interaction rounds without a conclusive diagnosis, the current specialist team is prompted to generate the best possible diagnosis with the available information, thereby concluding the session. All prompts are detailed in Appendix\ref{app:doctor_open}.

\section{Experiments and Results}

\subsection{Experimental Setup}
We evaluate both of our proposed Doctor System and its single-agent variant in DynamiCare on the MIMIC-Patient dataset (500 patient records) using GPT-4.1 and GPT-4o-mini. In the single-agent setting, the multi-agent specialist team is disabled and only one agent performs diagnostics per round. 
GPT-4.1 is also used in the Patient System for answer generation. For comparison with external baselines, we test our model on the MEDIQ benchmark\cite{li2024mediq}, using all 140 samples from iCraftMD and 200 randomly selected cases from iMedQA, following MEDIQ’s interactive multiple-choice setting and using GPT-4.1 for all runs.

\subsection{Experimental Results}
% justify multi-agent design: more powerful than single agent

% Patient Agent(MIMIC dataset):\\
% multi-agent vs single (gpt-4.1, gpt-4o-mini)\\

% MEDIQ dataset: \\ 
% multi-agent vs MEDIQ (gpt-4.1)

\begin{table*}[t]
\centering
%\resizebox{\textwidth}{!}{
%{\small   %% Make the font smaller
\begin{tabular}{ccccccc}
\toprule
\multicolumn{1}{l}{Agent} & \multicolumn{1}{c}{ GPT Version} & \multicolumn{1}{c}{Hit@5} & \multicolumn{1}{c}{Hit@10} & \multicolumn{1}{c}{Rec@5} & \multicolumn{1}{c}{Rec@10} & \multicolumn{1}{c}{Ave-Q}\\
\midrule
\multirow{2}{*}{\begin{tabular}[c]{@{}l@{}}{{Multi}}\end{tabular}} & GPT-4.1 & \textbf{63.4}  & \textbf{71.6} & \textbf{43.2} & \textbf{58.8} & 7.55\\
& GPT-4o-mini & 51.4  & 58.6 & 30.2 & 43.2 & 2.78\\
\midrule
\multirow{2}{*}{\begin{tabular}[c]{@{}l@{}}{{Single}}\end{tabular}} & GPT-4.1 & 58.0 & 63.2  & 31.0 & 41.7 & 3.83\\
& GPT-4o-mini & 47.8  & 54.8 & 24.9 & 31.8 & 0.74\\
\bottomrule
\end{tabular}
\caption{Performance comparison between the proposed multi-agent Doctor System and a single-agent variant across two LLMs (GPT-4.1 and GPT-4o-mini). Metrics include Hit@K, Rec@K, and Ave-Q (average number of questions asked).}
\label{table: performance}
\end{table*}

Table \ref{table: performance} presents the performance of our proposed Doctor System, including its multi- and single-agent variants (with the specialist team fixed to the single-agent version). The backbone models use GPT-4.1 and GPT-4o-mini. The results demonstrate the effectiveness of the dynamic multi-agent setting across multiple metrics and model variants.

We then evaluate the results using the top-k hit rate (Hit@5, Hit@10)~\footnote{Hit@K measures whether at least one of the matched (ground truth) items appears in the top K results returned by the model.} and recall (Rec@5, Rec@10)~\footnote{Recall@K measures the proportion of all matched items that are found in the top K results.}, which measure the system's ability to rank correct diagnoses among its predictions. As the doctor agent is prompted to return a list of up to 10 likely diagnoses, and the ground truth typically includes between 1 to 5 correct labels, Hit@K metrics help assess how well the system captures correct diagnoses within a ranked list, whereas Rec@K offers a perspective on how many ground truth diagnoses are successfully identified.

To standardize diagnosis terminology and enable consistent comparison, we map all predicted diagnoses to ICD-9 codes using the BioPortal API~\cite{whetzel2011bioportal}. We consider a prediction correct if the first three digits of the predicted ICD-9 code match any of the ground truth codes. This reflects a clinically meaningful level of accuracy, as the first three digits of an ICD-9 code typically represent the high-level diagnostic category.~\cite{rajkomar2018scalable,shi2017towards}. It is worth noting that, even under this relatively relaxed criterion, the classification task remains extremely challenging, with over 1,000 possible 3-digit ICD-9 code groups in the prediction space.

From Table \ref{table: performance}, we observe that the proposed dynamic multi-agent system consistently outperforms the single-agent variant across all metrics and LLM models. These improvements indicate that dynamically adjusting the specialist team based on patient complexity—as implemented in the multi-agent setting—enhances diagnostic performance, especially in our setting involving open-ended medical diagnosis with rich and often ambiguous patient information.

Additionally, model quality plays a significant role. GPT-4.1 consistently outperforms GPT-4o-mini across all settings, emphasizing the importance of a strong language model backbone in achieving accurate medical reasoning. This is further supported by Ave-Q, the average number of questions asked per patient case, where we observe more questions asked with stronger models and more complex settings, reflecting a more interactive diagnostic process.

\subsection{Cross Comparison with MEDIQ}
\begin{table}[h]
\centering
%\resizebox{\textwidth}{!}{
%{\small   %% Make the font smaller
\begin{tabular}{lcc}
\toprule
\multicolumn{1}{l}{Agent} & \multicolumn{1}{c}{ Dataset} & \multicolumn{1}{c}{Accuracy} \\
\midrule
\multirow{2}{*}{\begin{tabular}[l]{@{}l@{}}{{MedIQ}}\end{tabular}} & iMedQA & 67.0 \\
& iCraftMD & 72.1\\
\midrule
\multirow{2}{*}{\begin{tabular}[l]{@{}l@{}}{{DynamiCare}}\end{tabular}} & iMedQA & \textbf{92.0} \\
& iCraftMD & \textbf{96.4}\\
\bottomrule
\end{tabular}
\caption{Accuracy comparison on multiple-choice question answering tasks from the MEDIQ benchmark, across two datasets (iMedQA and iCraftMD). Our dynamic multi-agent system (DyamiCare) outperforms the MedIQ baseline under the same GPT-4.1 setup.}
\label{table: compare}
\end{table}

To further validate the effectiveness of our dynamic multi-agent setting, we conduct additional evaluations on the MEDIQ interactive benchmark\cite{li2024mediq}, which includes two interactive QA tasks: iMedQA and iCraftMD. These tasks are framed as multiple-choice question answering (MCQ), which is relatively simpler than our open-ended diagnosis generation task.
For evaluation, we use the full iCraftMD dataset (140 samples) and randomly select 200 cases from iMedQA. Prompt templates for our Doctor System are appropriately adapted to match the multiple-choice format, all experiments use GPT-4.1.
As shown in Table \ref{table: compare}, our Doctor System significantly outperforms the best MEDIQ baseline in both datasets, further demonstrating the capability of our model.
\subsection{Patient System Evaluation}
% Reliability: Select some question and answer pairs for evaluation, are they truthful to the original data? Are they related to the question?
% 1. randomly selected 100 Q&A pairs from visit logs(3775 intotal)
% 2. For each question-answer pair, annotate along two dimensions:
% Truthfulness (Factuality): Is the answer consistent with the structured data (e.g., JSON records, MIMIC-III tables)?
% Scale: 0 = incorrect, 1 = partially correct, 2 = fully correct
% Relevance: Does the answer directly and sufficiently address the doctor's question?
% Scale: 0 = irrelevant, 1 = somewhat related, 2 = highly relevant
% 3. have 3 medical students do manual evaluation
\begin{table}[h]
\begin{tabular}{lllll}
\toprule
Metric       & A    & B    & C    & Average \\ \midrule
Truthfulness & 1.99 & 1.95 & 1.92 & 1.95    \\
Relevance    & 1.84 & 1.75 & 1.79 & 1.79    \\ \bottomrule
\end{tabular}
\caption{Manual evaluation of simulated patient responses across 100 patient sessions. Each response was rated by three annotators on Truthfulness and Relevance using a 3-point scale (0–2). The table reports the average scores per annotator and the overall mean across annotators.}
\label{tab:quality}
\end{table}
To assess the quality of the responses generated by our Patient System, we conduct a manual evaluation of 100 randomly selected patients' question-answer history (multi-agent setting, GPT-4.1). Each Q\&A log is independently annotated by three medical students along two dimensions: Truthfulness and Relevance. 
\begin{itemize}
    \item \textbf{Truthfulness} measures whether the patient's response is consistent with his/her JSON record, using a 3-point scale: 0 = incorrect, 1 = partially correct, 2 = fully correct.
    \item \textbf{Relevance} assesses whether the answer directly and adequately responds to the doctor's question, also using a 3-point scale: 0 = irrelevant, 1 = somewhat related, 2 = highly relevant.
\end{itemize}
The scores from the three annotators were averaged across all questions per patient and reported separately for each annotator (A, B, C), along with the overall mean as shown in Table \ref{tab:quality}. This evaluation provides a quantitative estimate of the reliability and contextual appropriateness of the patient agent’s responses, ensuring that it supports trustworthy and contextually appropriate interactions throughout the diagnostic process.

\subsection{Disease Case Study}
% does different disease type affect prediction accuracy?(some diseases may be harder to diagnosis)
\begin{table*}[t]
\centering
\resizebox{\textwidth}{!}{
\begin{tabular}{llccc}
\toprule
ICD-9 codes & Definition                                                   & Hit@5 & Hit@10 & Sample Size \\ \midrule
630-679     & complications of pregnancy, childbirth,   and the puerperium & \cellcolor{green!20} 100   & \cellcolor{green!20} 100    & 1           \\
760-779     & certain conditions originating in the   perinatal period     & \cellcolor{green!20} 85.71 & \cellcolor{green!20} 85.71  & 7           \\
710-739       & diseases of the musculoskeletal system   and connective tissue          & \cellcolor{green!20} 76.00 & \cellcolor{green!20} 76.00 & 25  \\
390-459     & diseases of the circulatory system                           & \cellcolor{green!20} 74.09 & \cellcolor{green!20} 81.73  & 301         \\
740-759     & congenital anomalies                                         & \cellcolor{green!20} 69.23 & \cellcolor{green!20} 76.92  & 26          \\
240-279       & endocrine, nutritional and metabolic   diseases, and immunity disorders & \cellcolor{green!20} 67.27 & \cellcolor{green!20} 77.58 & 165 \\
520-579     & diseases of the digestive system                             & \cellcolor{green!20} 63.77 & \cellcolor{green!20} 72.46  & 69          \\
460-519     & diseases of the respiratory system                           & 62.90 & \cellcolor{green!20} 72.58  & 62          \\
320-389     & diseases of the nervous system and sense   organs            & 61.22 & 71.43  & 49          \\
290-319     & mental disorders                                             & 57.97 & 69.57  & 69          \\
280-289     & diseases of the blood and blood-forming   organs             & 55.56 & \cellcolor{green!20} 75.00  & 36          \\
800-999     & injury and poisoning                                         & 50.56 & 59.55  & 89          \\
580-629     & diseases of the genitourinary system                         & 50.00 & 66.67  & 12          \\
680-709     & diseases of the skin and subcutaneous   tissue               & 50.00 & 50.00  & 6           \\
780-799     & symptoms, signs, and ill-defined   conditions                & 48.78 & 68.29  & 41          \\
E and V codes & external causes of injury and   supplemental classification             & 48.28 & 57.93 & 145 \\
140-239     & neoplasms                                                    & 32.00 & 40.00  & 50          \\
001–139     & infectious and parasitic diseases                            & -     & -      & 0           \\ \bottomrule
\end{tabular}}
\caption{Diagnostic accuracy across ICD-9 code categories. Each row represents a high-level disease group, along with its corresponding accuracy and sample size. Categories with accuracy above the overall mean (Hit@5: 63.4\%, Hit@10: 71.6\%) are highlighted in green. Sample size indicates the number of patient–diagnosis instances assigned to each ICD-9 category; since each patient may have multiple codes, the total exceeds the number of unique patients.}
\label{tab:casestudy}
\end{table*}

We observe substantial variation in diagnostic performance across ICD-9 classes, with top-5 accuracy ranging from 32\% to 100\% (mean: 63.4\%), and top-10 accuracy from 40\% to 100\% (mean: 71.6\%). Excluding classes with insufficient sample size, high-performing categories include diagnoses like musculoskeletal diseases and circulatory system disorders, whereas lower performance is seen in neoplasms, external causes of injury, and symptoms or ill-defined conditions.

These differences appear to stem from three main factors. First, \textbf{case complexity} plays a key role \cite {khan2024comparison, mcduff2025towards}. Conditions with well-defined, localized symptoms—such as musculoskeletal or circulatory disorders—tend to involve fewer comorbidities and more predictable clinical patterns, enabling the model to identify the correct diagnosis with fewer reasoning steps. In contrast, complex conditions like neoplasms or non-specific symptom clusters often involve overlapping or ambiguous presentations that require integration of contextual or longitudinal information, making accurate diagnosis considerably more challenging within a constrained Q\&A framework.

Second, \textbf{disease prevalence and representation} may influence GPT's performance \cite{sandmann2024systematic}. Although the system is not fine-tuned, its base knowledge reflects conditions that are more frequently discussed in medical literature and clinical practice. As a result, the model tends to perform better on prevalent diseases with well-defined patterns (e.g., cardiovascular and circulatory diseases). 

Finally, the \textbf{disease-specific nature of Q\&A interactions} influences diagnostic success. Diseases with clear symptomatology allow for efficient question targeting, enabling quick narrowing the differential diagnosis. Conversely, vague or multi-system conditions—such as those classified under neoplasms or ill-defined symptoms—require broader questioning and yield more diffuse information, often resulting in longer differential lists where the correct diagnosis may be deprioritized. Additionally, categories requiring contextual or social information (e.g., external causes of injury) are more likely to be missed when such details are not explicitly elicited.

\section{Discussion}
This study introduces a dynamic multi-agent framework for clinical decision-making, grounded in a realistic simulation of the diagnostic process. Unlike prior works that rely on single-turn or static multi-agent setups, our system dynamically adapts its specialist composition in response to evolving patient information. Experiments on the MIMIC-derived open-ended diagnosis benchmark demonstrate that our model consistently outperforms a single-agent variant, particularly when reasoning over complex and ambiguous cases. Moreover, our patient system was shown to produce responses that are both truthful and relevant, supporting a coherent and trustworthy dialogue process.  

% Limitation and future work:\\
% 1. now we only have text and tabular data, can include more data modalities in the future (eg. image, multiomics)\\
% 2. patients can have preferences on their information(more likely to include important info, even unasked)\\
% 3. more techniques can be applied to the doctor system: RAG for example
% Despite the promising results, several limitations remain:
\section{Limitations}
While our approach shows promising results, several limitations remain, opening opportunities for future research and improvement.
First, our system currently operates on textual and tabular data only. Incorporating other clinically significant modalities such as medical imaging, genomics, or sensor data could enable richer and more accurate diagnostic reasoning.
Second, real patients may volunteer important information even if not directly asked. Future versions of the patient system could be designed to simulate such proactive behavior, improving the realism of interactions.
Third, while our dynamic framework already improves reasoning, further enhancement may come from integrating retrieval-augmented generation (RAG), external medical knowledge bases, or modular expert components fine-tuned for specific specialties.

\bibliography{refs/main}

\appendix
\clearpage
\onecolumn
\section{Patient System}
\label{app: patient}

%%%%%%%%Patient JSON%%%%%%%%

% \begin{table}[ht]
% \centering
% \begin{tabular}{@{}l p{10cm}@{}}
% \toprule
% \textbf{Data Type} & \textbf{Category} \\
% \midrule
% Admission Info & Details of hospital admission: admission/discharge time, type, and reason. \\
% Demographics & Basic patient info: age, sex, ethnicity, language, marital status. \\
% Diagnoses & Diagnosis codes and descriptions (ICD-9/10), including primary and secondary. \\
% Prescription & Medications prescribed: drug names, dosages, routes, and durations. \\
% Procedure & Clinical procedures performed, with codes and timestamps. \\
% ECG & Electrocardiogram reports and interpretations. \\
% Echo & Echocardiogram results, including measurements like ejection fraction. \\
% Radiology & Imaging reports such as X-ray, CT, MRI with impressions. \\
% Discharge Summary & Final summary of the hospital stay including condition, treatment, and plans. \\
% Chart Data & Time-series data on vitals, inputs/outputs, and nursing observations. \\
% Lab Data & Laboratory test results with values and timestamps. \\
% \bottomrule
% \end{tabular}
% \caption{Overview of Patient Information Categories}
% \label{table: patient info}
% \end{table}

\begin{tcolorbox}[colback=yellow!10, colframe=gray, title=Patient JSON (example)] \label{app:patient json}
\begin{lstlisting}[language=Python]
"Patient": {
    "Admission_info": {
      "patient_id": ****,
      "admission_id": ****,
      "admission_diagnosis": "arrhythmia"
    },
    "Demographics": {
      "insurance": "private",
      "language": "engl",
      "marital_status": "married",
      "ethnicity": "white",
      "gender": "M",
      "age": 60
    },
    "Diagnoses": [[
        "4019",
        "Hypertension NOS",
        "Unspecified essential hypertension"
      ],...],
    "Prescription": [
      "Sodium Chloride 0.9%  Flush",
      "Lisinopril",
      "Heparin",...],
    "Introduction": "Hi, I'm a 60-year-old male. I was referred here by my clinic ...my doctor was concerned about a possible arrhythmia.",
    "ECG": [[
        "2105-03-03",
        "Atrial pacing and A-V conduction which is new compared to previous tracings."
      ],...],
    "Radiology": [{
        "time": "2105-03-03",
        "part": "CHEST (PA & LAT)",
        "medical condition": "60 year old man with new dual chamber",
        "ppm reason for this examination": "Evaluate lead position",
        "final report history": "Pacemaker placement.",
        "findings": "In comparison with the study of , there has been placement of ... other acute cardiopulmonary disease."
      }...],
    "Allergies": "No Known Allergies / Adverse Drug Reactions",
    "Chief Complaint": "Fatigue, lightheadedness, bradycardia, sinus pauses",
    "Major Surgical or Invasive Procedure": "Pacemaker placement (St. Medical Accent PM2210 dual chamber pacemaker)",
    "Physical Exam": { "Admission": {
        "VS": "T=98.0 BP=158/91 HR=61 RR=18 O2 sat=95",
        "General": "WDWN M in NAD. Oriented x3. Mood, affect appropriate. Fit",
        "HEENT": "NCAT. Sclera anicteric. PERRL, EOMI. Conjunctiva were pink, no pallor or cyanosis of the oral mucosa. No xanthalesma.",
        "Neck": "Supple with JVP below clavicle at 90 degrees.",
        "Cardiac": "PMI located in 5th intercostal space, midclavicular line. RR, normal S1, S2. No m/r/g. No thrills, lifts. No S3 or S4.",
        "Lungs": "No chest wall deformities, scoliosis or kyphosis. Resp were unlabored, no accessory muscle use. CTAB, no crackles, wheezes or rhonchi.",
        "Abdomen": "Soft, NTND. No HSM or tenderness. Abd aorta not enlarged by palpation. No abdominial bruits.",
        "Extremities": "No c/c/e. No femoral bruits.",
        "Skin": "No stasis dermatitis, ulcers, scars, or xanthomas.",
        "Pulses": {
          "Right": "Carotid 2+ Femoral 2+ Popliteal 2+ DP 2+ PT 2+",
          "Left": "Carotid 2+ Femoral 2+ Popliteal 2+ DP 2+ PT 2+"}
      },...},
    "Respiratory": { "O2 saturation pulseoxymetry": [ ["2105-02-28 03:15:00", "97.0 %"],...],...}
\end{lstlisting}
\end{tcolorbox}

%%%%%%%%Keyword Mapping%%%%%%%%
\begin{tcolorbox}[colback=yellow!10, colframe=gray, title=Keyword Mapping Dictionary (example)] \label{app:key-words}
\begin{lstlisting}[language=Python]
keyword_mapping = {
    # Demographics
    ('age',): ('Demographics', 'age'),
    ('language', 'english', 'spanish'): ('Demographics', 'language'),
    ('religion', 'religious'): ('Demographics', 'religion'),
    ('marital', 'married', 'single', 'divorced', 'widowed'): ('Demographics', 'marital_status'),
    ('gender', 'sex'): ('Demographics', 'gender'),
    ('insurance',): ('Demographics', 'insurance'),
    ('ethnicity',): ('Demographics', 'ethnicity'),
    # Medications
    ('prescription', 'medications', 'medication', 'drugs'): ('Prescription', None),
    ('admission medications', 'initial meds'): ('Medications on Admission', None),
    # Procedures
    ('procedure', 'surgery', 'operation'): [('Procedure', None), ('Major Surgical or Invasive Procedure', None)],
    # Imaging and reports
    ('electrocardiogram', 'ecg'): ('ECG', None),
    ('echocardiogram', 'echo'): ('Echo', None),
    ('radiology', 'x-ray', 'ct', 'mri', 'imaging'): ('Radiology', None),
    # History
    ('hpi', 'present illness', 'history of present illness'): ('History of Present Illness', None),
    ('past medical', 'pmh', 'past medical history'): ('Past Medical History', None),
    ('family history',): ('Family History', None),
    ('social history', 'drinking', 'smoking', 'drug use', 'tobacco', 'alcohol'): ('Social History', None),
    # Allergies
    ('allergy', 'allergies', 'allergic'): ('Allergies', None),
    # Physical exam
    ('heent', 'head', 'eyes', 'ears', 'nose', 'throat'): ('Physical Exam.Admission', 'HEENT'),
    ('physical exam',): ('Physical Exam.Admission', None),
    ...}
\end{lstlisting}
\end{tcolorbox}

\section{Doctor Agent on MIMIC Open-Ended Diagnosis}
\label{app:doctor_open}
%%%%%%%%Triage Prompt%%%%%%%%
\begin{tcolorbox}[colback=yellow!10, colframe=gray, title=Initial Specialist Triage Prompt]
\begin{lstlisting}[language=TeX]
You are a general practitioner triaging a new patient. Based on the patient's initial admission information, recommend one or more medical specialists to consult.
The maximum number of specialists is 5.
Return your answer in the following JSON format only:
{
"RATIONALE": "<short justification>",
"SUGGEST_SPECIALISTS": [<list of specialists>] 
}
\end{lstlisting}
\end{tcolorbox}

%%%%%%%%Specialist Update Prompt%%%%%%%%
\begin{tcolorbox}[colback=yellow!10, colframe=gray, title=Central Agent Coordination Prompt]
\begin{lstlisting}[language=TeX]
You are a central medical coordinator overseeing a diagnostic team. 
Based on the current patient case and the specialists already involved, decide if additional experts are needed or if any can be removed.
Only suggest adding new specialists if there's missing domain knowledge. Only suggest removing specialists if their role has been fully covered or no longer needed.
The maximum number of specialists is 5.
Respond in this JSON format only:
{
"ADD": [<specialists to add>],
"REMOVE": [<specialists to remove>],
"UPDATED_LIST": [<updated specialists>],
"RATIONALE": "<short justification>"
}
\end{lstlisting}
\end{tcolorbox}

%%%%%%%%Confidence Rating Prompt%%%%%%%%
\begin{tcolorbox}[colback=yellow!10, colframe=gray, title=Specialist Confidence Rating Prompt]
\begin{lstlisting}[language=TeX]
You are a {spec}. Based on the current patient case, decide whether you are confident enough to make a diagnosis or whether more information is needed.

Choose between the following ratings: 
"Very Confident"- The diagnosis is strongly supported by current information, and no major uncertainties remain.
"Somewhat Confident"- The diagnosis is likely given the evidence, but a bit more information would increase certainty.
"Neither Confident or Unconfident"- Some clues suggest a possible diagnosis, but key details are still missing. 
"Somewhat Unconfident"- Several diagnoses remain plausible; more data is needed to narrow them down.
"Very Unconfident"- There is too little evidence to form a reasonable diagnostic opinion.

Respond in the following format only: 
DECISION: chosen rating from the above list.
\end{lstlisting}
\end{tcolorbox}

%%%%%%%%Solo Specialist Decision Prompt%%%%%%%%
\begin{tcolorbox}[colback=yellow!10, colframe=gray, title=Solo Specialist Diagnosis or Question Prompt]
\begin{lstlisting}[language=TeX]
You are a {spec}. Based on the current patient case, list the top 10 most likely diagnoses for this patient.
Only include the **diagnosis name** in the list.
Respond in this JSON format only:
{
"RESPONSE_TYPE": "diagnosis",
"RESPONSE_CONTENT": "[<your diagnosis list>]",
"RATIONALE": "<brief justification>"
}
\end{lstlisting}
\end{tcolorbox}

%%%%%%%%Follow-up Question Prompt%%%%%%%%
\begin{tcolorbox}[colback=yellow!10, colframe=gray, title=Solo Specialist Follow-Up Question Prompt]
\begin{lstlisting}[language=TeX]
You are a {spec}. Based on your medical expertise and the current information, propose the most important next question that would help you narrow down or confirm a diagnosis.
The question should be specific and relevant to the case.
Do not repeat any questions from the previous conversation log or ask about information already provided.
Avoid asking about topics already answered with "I don't know" or "not in chart"
If referencing labs, vitals, ECG, radiology, etc. - which may have multiple time points - be clear about the desired time window.
Respond in this JSON format only:
{
"RESPONSE_TYPE": "question",
"RESPONSE_CONTENT": "<your follow-up question>",
"RATIONALE": "<brief justification>"
}
\end{lstlisting}
\end{tcolorbox}

%%%%%%%%Collaborative Decision Prompt%%%%%%%%
\begin{tcolorbox}[colback=yellow!10, colframe=gray, title=Collaborative Decision Proposal Prompt]
\begin{lstlisting}[language=TeX]
You are a {spec}. Based on your medical expertise and the current information, respond with either:
- A list of the top 10 most likely diagnoses (if you believe you can make a clinical judgment now), or
- A follow-up question (if you believe more information is needed to proceed)

Only include the **diagnosis name** in the list.
The question should be specific and relevant to the case.
Do not repeat any questions from the previous conversation log or ask about information already provided.
Avoid asking about topics already answered with "I don't know" or "not in chart"
If referencing labs, vitals, ECG, radiology, etc. - which may have **multiple time points** - be clear about the **desired time window**.

Respond in this JSON format only:
{
"RESPONSE_TYPE": "<diagnosis | question>",
"RESPONSE_CONTENT": "<your diagnosis list or follow-up question>",
"CONFIDENCE": "<1-5>",  // 5 = Very confident on the appropriateness of the response,
"RATIONALE": "<brief justification>"
}
\end{lstlisting}
\end{tcolorbox}

%%%%%%%%Voting Prompt%%%%%%%%
\begin{tcolorbox}[colback=yellow!10, colframe=gray, title=Voting Prompt]
\begin{lstlisting}[language=TeX]
You are a {voter['SPECIALIST']}. Another specialist ({candidate['SPECIALIST']}) proposed the following:

They suggested to proceed with a **{candidate['RESPONSE_TYPE']}**, which is:
"{candidate['RESPONSE_CONTENT']}"

Their rationale is:
"{candidate['RATIONALE']}"

Based on the current patient case, decide if you agree with the proposed next step.
Respond with ONLY one of the following options:
- AGREE
- DISAGREE
\end{lstlisting}
\end{tcolorbox}
\clearpage
\section{Potential Risk}
While DynamiCare is designed for research and educational purposes, there are potential risks associated with the misuse or misinterpretation of its outputs. The system simulates clinical reasoning using language models that, despite strong performance, may generate incorrect, incomplete, or misleading responses if taken at face value. If such models were applied in real-world medical settings without appropriate oversight, they could lead to harmful diagnostic decisions or delayed treatments. Additionally, since MIMIC-Patient is derived from real patient data (via the de-identified MIMIC-III database), there is a need to ensure responsible data handling and prevent re-identification risks, even though direct identifiers have been removed. Finally, overreliance on open-ended AI-generated diagnoses could inadvertently reinforce biases or amplify gaps in model training data. We emphasize that our framework is not intended for clinical deployment and should be used strictly in controlled, academic environments.

\end{document}